%% file: main.tex
\documentclass{article}

\usepackage{microtype}
\usepackage{graphicx}
\usepackage{subfigure}
\usepackage{booktabs} %

\usepackage{hyperref}

\usepackage[arxiv]{icml2025}

\usepackage{amsmath}
\usepackage{amssymb}
\usepackage{mathtools}
\usepackage{amsthm}

\usepackage[capitalize,noabbrev]{cleveref}

\usepackage{multirow}
\usepackage{makecell}
\usepackage{float}
\usepackage{color,xcolor}
\usepackage{colortbl}
\usepackage{caption} 
\theoremstyle{plain}

\theoremstyle{definition}

\theoremstyle{remark}

\usepackage[textsize=tiny]{todonotes}

\newcommand{\boldparagraph}[1]{\vspace{0.05cm}\noindent{\bf #1}}
\newcommand{\name}{mmMamba}

\def\eg{\emph{e.g.}} 
\def\ie{\emph{i.e.}}

\icmltitlerunning{Multimodal Mamba: Decoder-only Multimodal State Space Model via Quadratic to Linear Distillation}

\begin{document}

\twocolumn[
\icmltitle{Multimodal Mamba: Decoder-only Multimodal State Space Model via Quadratic to Linear Distillation}

\icmlsetsymbol{equal}{*}

\begin{icmlauthorlist}
\icmlauthor{Bencheng Liao}{equal,aia,eic}
\icmlauthor{Hongyuan Tao}{equal,eic}
\icmlauthor{Qian Zhang}{horizon}
\icmlauthor{Tianheng Cheng}{eic}
\icmlauthor{Yingyue Li}{eic}
\icmlauthor{Haoran Yin}{horizon}
\icmlauthor{Wenyu Liu}{eic}
\icmlauthor{Xinggang Wang}{eic}
\end{icmlauthorlist}

\icmlaffiliation{aia}{Institute of Artificial Intelligence, Huazhong University of Science \& Technology}
\icmlaffiliation{eic}{School of EIC, Huazhong University of Science \& Technology}
\icmlaffiliation{horizon}{Horizon Robotics}

\icmlcorrespondingauthor{Xinggang Wang}{xgwang@hust.edu.cn}

\icmlkeywords{Machine Learning, ICML}

\vskip 0.3in
{%

\vspace{-0.5cm}
\begin{center}
    \captionsetup{type=figure}
    \includegraphics[width=0.9\textwidth]{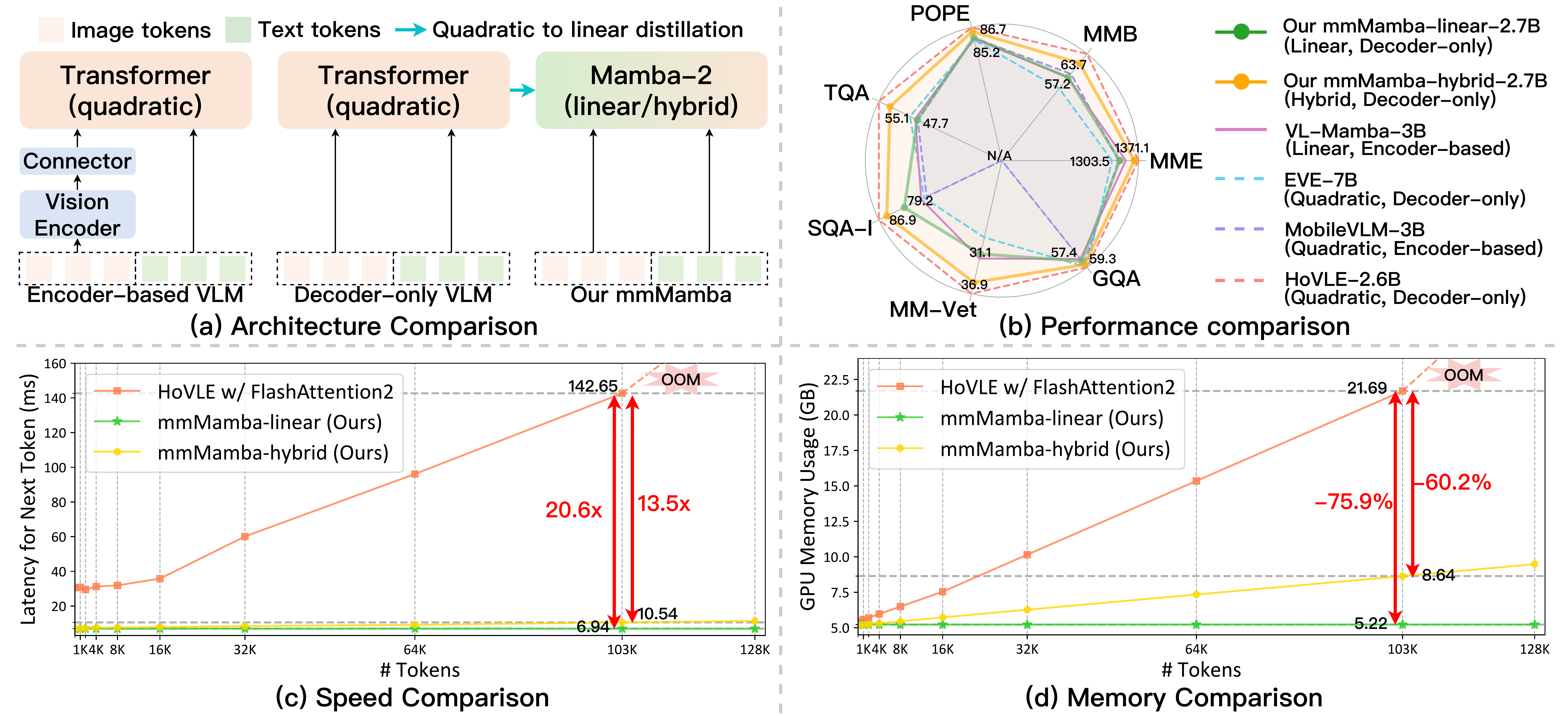}
    \vspace{-.1 in}
    \captionof{figure}{
        \textbf{Comprehensive comparison of \name{}.} 
        \textbf{(a)} Our \name{} can build linear-complexity and hybrid decoder-only VLM by distilling the knowledge in Transformer to Mamba-2. 
        \textbf{(b)} By distilling from the quadratic-complexity decoder-only VLM HoVLE,  our \name{}-linear achieves competitive performance against existing linear and quadratic-complexity VLMs with fewer parameters (\eg, 2$\times$ fewer than EVE-7B), while \name{}-hybrid surpasses them across all benchmarks and approaches the teacher model HoVLE's performance.
        \textbf{(c)-(d)} We compare the speed and memory of \name{}-linear and \name{}-hybrid with the teacher model HoVLE on the same single NVIDIA 4090 GPU. \name{}-linear maintains consistently low latency and memory usage, while \name{}-hybrid's resource consumption scales significantly better than HoVLE. At 103K tokens, \name{}-linear demonstrates 20.6$\times$ speedup compared to HoVLE and saves 75.8\% GPU memory, while \name{}-hybrid achieves 13.5$\times$ speedup and saves 60.2\% GPU memory.
    }
    \label{fig:teaser}
\end{center} %
}
]

\printAffiliationsAndNotice{\icmlEqualContribution} %

\begin{abstract}
Recent Multimodal Large Language Models (MLLMs) have achieved remarkable performance but face deployment challenges due to their quadratic computational complexity, growing Key-Value cache requirements, and reliance on separate vision encoders. We propose \name{}, a framework for developing linear-complexity native multimodal state space models through progressive distillation from existing MLLMs using moderate academic computational resources. Our approach enables the direct conversion of trained decoder-only MLLMs to linear-complexity architectures without requiring pre-trained RNN-based LLM or vision encoders. We propose an seeding strategy to carve Mamba from trained Transformer and a three-stage distillation recipe, which can effectively transfer the knowledge from Transformer to Mamba while preserving multimodal capabilities. Our method also supports flexible hybrid architectures that combine Transformer and Mamba layers for customizable efficiency-performance trade-offs.
Distilled from the Transformer-based decoder-only HoVLE, \name{}-linear achieves competitive performance against existing linear and quadratic-complexity VLMs, while \name{}-hybrid further improves performance significantly, approaching HoVLE's capabilities. At 103K tokens, \name{}-linear demonstrates 20.6$\times$ speedup and 75.8\% GPU memory reduction compared to HoVLE, while \name{}-hybrid achieves 13.5$\times$ speedup and 60.2\% memory savings.
Code and models are released at \url{https://github.com/hustvl/mmMamba}
\end{abstract}

\input{secs/1_intro}

\input{secs/2_related}

\input{secs/3_method}

\input{secs/4_exp}

\input{secs/5_conclusion}

\bibliography{example_paper}
\bibliographystyle{icml2025}

\end{document}

%% file: secs/1_intro.tex
\section{Introduction}
Recent advances in Large Language Models (LLMs)~\cite{gpt3, gpt4, llama, llama2, llama3, qwenllm, mistral, deepseek, phi2} have catalyzed significant research interest in expanding their capabilities beyond text to encompass multimodal understanding, particularly in processing both visual and textual information simultaneously. This expansion has given rise to Multimodal Large Language Models (MLLMs), with Vision Language Models (VLMs) emerging as a prominent subset. Notable examples such as LLaVA~\cite{llava}, BLIP~\cite{blip}, Qwen-VL~\cite{qwenvl}, InternVL~\cite{internvl}, and Monkey~\cite{monkey} have demonstrated remarkable success in enhancing LLMs' visual comprehension capabilities through the integration of pre-trained vision encoders and specialized connectors that bridge the modality gap between vision and language. 

While these encoder-based compositional VLMs have achieved state-of-the-art (SOTA) performance and established themselves as the de-facto paradigm, they face two critical limitations. 
First, processing long contexts becomes prohibitively expensive due to the quadratic computational complexity and linearly growing Key-Value (KV) cache with respect to sequence length. 
This limitation becomes particularly problematic given the increasing demand for long chain-of-thought reasoning~\cite{s1,deepseekr1,k1p5,llavacot} and high-resolution image/video understanding~\cite{videollm,hrvlm,vim,vig,dig}. Second, their heterogeneous architecture heavily relies on pre-trained vision encoders, introducing significant complexity in both training procedures and deployment scenarios~\cite{solo}.

Current research efforts to address these limitations have followed two distinct paths. One approach focuses on developing linear-complexity VLMs by adhering to the conventional encoder-based recipe, which requires both pre-trained vision encoders and pre-trained linear-complexity language models~\cite{visualrwkv,vlmamba}. The alternative approach aims to enhance decoder-only VLMs through increased model scale and expanded training datasets, achieving performance competitive with encoder-based counterparts~\cite{fuyu,eve,hovle,chameleon,emu3}.

Despite these advances, the development of linear-complexity decoder-only MLLMs remains an understudied yet critical challenge. Addressing this gap holds substantial value for three key reasons: (1) \textit{Unified multimodal understanding}: Such models could seamlessly integrate multimodal reasoning within a single architecture, eliminating the need for heterogeneous, modality-specific frameworks. (2) \textit{Practical efficiency}: Linear-complexity models inherently reduce computational demands during both training and inference, lowering costs and enabling deployment on resource-constrained edge devices. (3) \textit{Untapped Potential}: While recent linear-time models like Mamba-2 demonstrate high text-processing capabilities, their ability to handle multimodal tasks—particularly in cross-modal alignment and reasoning—remains largely unexplored. The research of linear-complexity decoder-only MLLMs could unlock scalable, cost-effective multimodal systems without sacrificing performance.

A straightforward solution is to synergize the recipe of decoder-only VLMs and linear-complexity encoder-based VLMs.
This integration requires a pre-trained linear-complexity LLM and performs image-caption alignment pre-training (PT) and supervised fine-tuning (SFT) using text instructions and image prompts. 
However, this integrated recipe faces two significant challenges:
(1) It demands the curation of different large-scale multimodal datasets for different purposes (\ie, PT and SFT) and requires substantial computational resources. 
(2) The overall performance is inherently limited by the capabilities of pre-trained linear-complexity LLMs, which consistently underperform mainstream SOTA Transformer-based LLMs in language understanding tasks.

In this paper, we propose a novel distillation-based recipe to develop linear-complexity decoder-only VLMs, which requires only moderate academic resources while circumventing the limitations of pre-trained linear-complexity LLMs.
Our method leverages the fundamental similarity between the Transformer attention mechanism and the Mamba-2 state space model (SSM) mechanism. We introduce an initialization scheme that enables direct parameter transfer from Transformer to Mamba-2 layers, effectively converting the attention mechanism into the SSM function while carefully initializing SSM-specific parameters to mimic attention behavior. This approach enables the direct transformation of pre-trained Transformer-based VLMs into linear-complexity Mamba-2-based VLMs without relying on underperforming pre-trained linear-complexity LLMs.
While this parameter inheritance and initialization strategy provides a promising starting point, the transformed Mamba-2-based VLM requires further distillation to recover robust multimodal conversation capabilities. To enhance alignment with the Transformer-based teacher VLM, we develop a three-stage progressive distillation strategy:
(1) Stage-1: we first train the SSM-specific parameters while freezing inherited parameters, and align layer-wise behavior using MSE distillation loss; 
(2) Stage-2: we then optimize complete Mamba-2 layer behavior by enabling the training of inherited Transformer parameters; 
(3) Stage-3: we finally perform complete model alignment using KL-divergence loss on final output logits to recover the teacher model's multimodal understanding capabilities through end-to-end distillation.

The proposed distillation recipe enables two distinct architectural variants: \name{}-linear, which converts all Transformer layers into Mamba-2 layers, achieving full linear complexity, and \name{}-hybrid, which strategically transforms fixed intervals of Transformer layers into Mamba-2 layers. The hybrid design systematically preserves Transformer layers at critical feature hierarchies while leveraging Mamba-2's linear complexity for the majority of computations, striking a balance between efficiency and capability. 
During the final end-to-end distillation stage, we can flexibly adjust the number of interleaved Transformer layers, enabling precise control over the computation-performance trade-off. This architectural flexibility makes our approach highly adaptable to diverse deployment scenarios, allowing optimization for specific computational constraints while maintaining desired performance.

By distilling from the recent Transformer-based decoder-only VLM, HoVLE, we demonstrate that \name{} achieves competitive performance across multiple vision-language benchmarks while significantly improving computational efficiency. Our pure Mamba-2-based linear-complexity variant, \name{}-linear, achieves comparable performance to existing quadratic/linear-complexity VLMs like Mobile-VLM-3B~\cite{mobilevlm}, VisualRWKV-3B~\cite{visualrwkv}, VL-Mamba-3B~\cite{vlmamba}
while eliminating the need for separate vision encoders. \name{}-pure also matches the performance of the previous SOTA Transformer-based decoder-only EVE-7B with 2$\times$ fewer parameters. The hybrid variant, \name-hybrid, significantly improves performance on all benchmarks compared to \name{}-pure, approaching the teacher model HoVLE. Notably, at the context length of 103K tokens, \name{}-linear demonstrates 20.6$\times$ speedup compared to HoVLE and saves 75.8\% GPU memory, while \name{}-hybrid achieves 13.5$\times$ speedup and saves 60.2\% GPU memory. These results and extensive ablation studies validate the effectiveness of our distillation recipe and highlight the potential for practical applications.

Our main contributions can be summarized as follows:
\begin{itemize}
\item We present a novel three-stage progressive distillation recipe for building native multimodal state space models without the reliance on underperforming pre-trained linear-complexity LLMs, enabling effective knowledge transfer from quadratic to linear architectures.
\item With the proposed distillation recipe, we propose the first decoder-only multimodal state space models that include two distinct architectural variants: \name{}-linear with purely linear complexity and \name{}-hybrid offering flexible performance-efficiency trade-offs.
\item Extensive experimental results demonstrate competitive performance with significantly improved computational efficiency across various vision-language tasks, achieving up to 20.6$\times$ speedup and 4.2$\times$ memory reduction for long sequence modeling on NVIDIA 4090 GPU.
\end{itemize}

%% file: secs/2_related.tex
\section{Related Work}
\boldparagraph{Decoder-only VLM.}
The remarkable success of Large Language Models (LLMs) has inspired the research community to extend their capabilities to multi-modal Vision-Language Models (VLMs).
While compositional encoder-based architectures~\cite{llava,internvl,monkey,blip}, leveraging pre-trained foundation vision encoders~\cite{eva, eva02, evaclip, sigclip} and additional connectors, have dominated the field. Recently, a pioneering work Fuyu-8B~\cite{fuyu} demonstrated that a single unified decoder-only Transformer can achieve competitive performance against encoder-based VLMs, offering an appealing alternative due to its architectural simplicity and deployment efficiency.
This breakthrough has sparked researchers' interest in decoder-only VLM. 
SOLO~\cite{solo} proposed a systematic training recipe tailored for decoder-only VLM by adapting pre-trained LLMs to vision-language tasks.
EVE~\cite{eve} advanced this approach by introducing vision-language pre-alignment and auxiliary visual representation supervision during fine-tuning to enhance the performance of decoder-only VLM.
To better preserve the inherited LLM's language capabilities, HoVLE~\cite{hovle} introduces an extra Transformer-based decoder-only holistic embedding module that aligns language and vision modalities before LLM processing multi-modal input tokens.
Despite these advances, existing decoder-only VLMs remain constrained by the quadratic computational complexity of Transformer architectures, resulting in substantial training and deployment costs. In contrast, our proposed \name{} addresses these limitations by converting Transformer layers to linear-complexity Mamba-2 layers through progressive distillation, enabling both pure linear and hybrid architectural variants.

\boldparagraph{Linear-complexity VLM.}
The development of linear-complexity RNN-based LLMs (\eg, Mamba~\cite{mamba}, Mamba-2~\cite{Mamba-2}, RWKV~\cite{rwkv}) has inspired increasing interest in addressing the quadratic complexity limitations of Transformer-based VLMs.
VL-Mamba~\cite{vlmamba} follows the recipe of LLaVA by incorporating a Vision Selective Scan connector with the pre-trained Mamba LLM.
Similarly, Cobra~\cite{cobra} enhances pre-trained Mamba LLM's visual capabilities by integrating DINOv2~\cite{dinov2} and SigLIP~\cite{sigclip} vision encoders.
ML-Mamba~\cite{mlmamba} introduces a Mamba-2 Scan Connector to process visual tokens between the pre-trained vision encoder and the pre-trained Mamba-2 LLM.
Instead of relying on Mamba, VisualRWKV~\cite{visualrwkv} leverages CLIP ViT-L/14~\cite{clip} as the vision encoder and a pre-trained RWKV LLM~\cite{rwkv,rwkv6} with a 2D image scanning mechanism for visual sequence processing.
However, the above works remain constrained by their reliance on pre-trained RNN-based LLMs and vision encoders, following the compositional encoder-based paradigm.
In contrast, our proposed \name{} eliminates the dependency on pre-trained RNN-based LLMs and vision encoders, and enables training a flexible hybrid architecture that interleaves Mamba with Transformer layers with minimal training cost. This capability enables customizable trade-offs between performance and efficiency, making it adaptable to diverse practical applications.

\boldparagraph{Transformer to RNN distillation.}
Instead of training RNN-based LLMs from scratch, recent studies propose to linearize the pre-trained Transformer-based LLMs into RNN-based LLMs through distillation, which can significantly reduce the training cost for building RNN-based LLMs.
\citet{kasai2021finetuning} pioneered this approach by using linear attention and initializing linear attention parameters using pre-trained LLM weights, exploiting the inherent similarities with Transformer's softmax attention.
\citet{zhang2024hedgehog} propose to add loss for matching softmax attention to approximate more closely the base transformer.   
\citet{mercat2024linearizing} advanced the field by replacing softmax attention with a linear RNN kernel coupled with a novel normalization strategy. 
Building upon these foundations, \citet{bick2024transformers}, \citet{wang2024mamba}, and \citet{zhang2024lolcats} developed multi-stage distillation approaches for more effective Transformer to RNN distillation.
Inspired by these advances, we extend this distillation paradigm to VLMs through the proposed novel multi-stage distillation strategy. Our approach first aligns the newly added parameters of the linearized LLM at each layer, followed by layer-wise distillation, and concludes with end-to-end distillation. This progressive pipeline ensures efficient transfer from quadratic knowledge to linear knowledge while maintaining performance.

%% file: secs/3_method.tex
\section{Preliminary}
\label{sec:preliminary}
We firstly give a brief background on quadratic-complexity sequence modeling Transformer and linear-complexity sequence modeling Mamba-2.
Given an input sequence $\mathbf{X} = \left[\boldsymbol{x}_1,\dots, \boldsymbol{x}_T\right]^\top \in \mathbb{R}^{T \times d}$, where $T$ is the sequence length and $d$ is the hidden dimension. The above two sequence modeling layers will compute the output sequence $\mathbf{Y} = \left[\boldsymbol{y}_1,\dots, \boldsymbol{y}_T\right]^\top \in \mathbb{R}^{T \times d}$.

\paragraph{Transformer}
The standard autoregressive Transformer used in LLM employs attention mechanism~\cite{attention} by interacting with all historical positions in the sequence, which is defined as:
\begin{align}
\begin{split}
    \boldsymbol{q}_t, \boldsymbol{k}_t, \boldsymbol{v}_t &= \boldsymbol{x}_t \boldsymbol{W}_Q, \boldsymbol{x}_t \boldsymbol{W}_K, \boldsymbol{x}_t \boldsymbol{W}_V, \\
    \boldsymbol{y}_t &= \frac{\sum_{i=1}^{t}\text{exp}(\boldsymbol{q}_t\boldsymbol{k}_i^\top)\boldsymbol{v}_i}{\sum_{i=1}^{t}\text{exp}(\boldsymbol{q}_t\boldsymbol{k}_i^\top)},
    \label{eq:attn}
\end{split}
\end{align}
where $\boldsymbol{W}_Q, \boldsymbol{W}_K, \boldsymbol{W}_V \in \mathbb{R}^{d \times d}$ are the learnable parameters. The current output token $\boldsymbol{o}_t$ is computed by performing attention over the growing sequence of historical keys $\left\{\boldsymbol{k}_i\right\}_{i=1}^{t}$ and values $\left\{\boldsymbol{v}_i\right\}_{i=1}^{t}$.

\paragraph{Mamba-2}
Instead of interacting with all historical positions, Mamba-2~\cite{Mamba-2} compresses the historical information into a fixed-size matrix-shaped hidden state, which is defined as:
\begin{align}
\begin{split}
    \boldsymbol{q}_t, \boldsymbol{k}_t, \boldsymbol{v}_t &= \boldsymbol{x}_t \boldsymbol{W}_Q, \boldsymbol{x}_t \boldsymbol{W}_K, \boldsymbol{x}_t \boldsymbol{W}_V, \\
    \gamma_t &= \text{exp}\left(-\text{softplus}(\boldsymbol{x}_t\boldsymbol{W}_\gamma)\text{exp}(a)\right), \\
    \mathbf{S}_t &= \gamma_t \mathbf{S}_{t-1} + \boldsymbol{v}_t \boldsymbol{k}_t^\top, \\
    \boldsymbol{y}_t &= \mathbf{S}_t \boldsymbol{q}_t,
    \label{eq:Mamba-2}
\end{split}
\end{align}
where $\boldsymbol{W}_Q, \boldsymbol{W}_K, \boldsymbol{W}_V \in \mathbb{R}^{d \times d}$, $\boldsymbol{W}_\gamma \in \mathbb{R}^{d \times 1}$ and $a \in \mathbb{R}$ are the learnable parameters. $\mathbf{S}_t$ is the fixed-size matrix-shaped hidden state, $\gamma_t$ is the data-dependent gating term to control the information flow by dynamically decaying the historical information $\mathbf{S}_{t-1}$.

\section{Method}
Our method consists of three key components. 
First, we detail the seeding strategy, which carves the Mamba-2 architecture from a pre-trained Transformer by inheriting parameters and carefully initializing the newly introduced SSM-specific parameters in Sec.~\ref{sec:init}. 
Building upon this seeding strategy, we present the proposed progressive distillation pipeline in Sec.~\ref{sec:stage1}, Sec.~\ref{sec:stage2} and Sec.~\ref{sec:stage3} to effectively transfer knowledge from Transformer to Mamba-2. 
With the designed distillation training recipe, we then instantiate two model variants in Sec.~\ref{sec:arch}: \name{}-linear using only Mamba-2 layers, and \name{}-hybrid incorporating interleaved Transformer and Mamba-2 layers.

\subsection{Seeding: Initialize Mamba-2 from Transformer}
\label{sec:init}
\begin{figure}
    \centering
    \includegraphics[width=\linewidth]{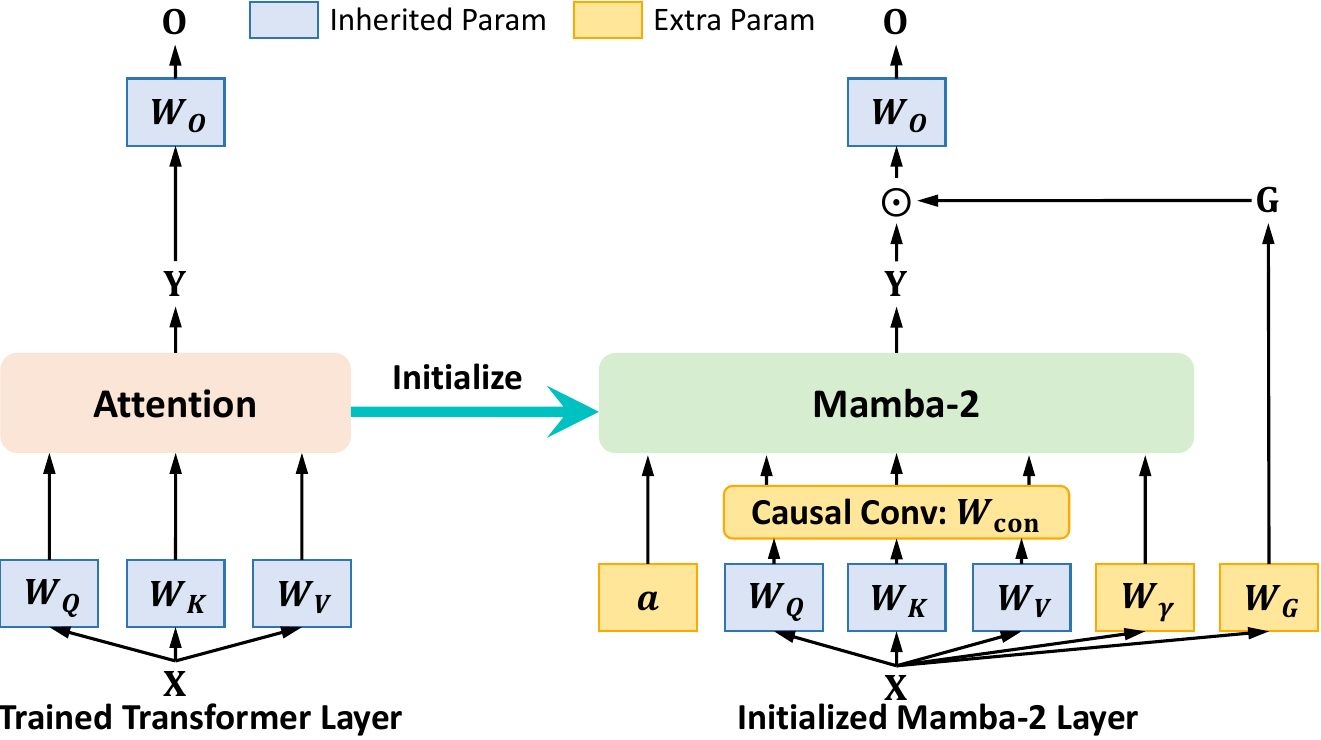}
    \caption{
        \textbf{Initialize Mamba-2 from Transformer.} By comparing the mechanism similarity in Sec.~\ref{sec:preliminary}, we directly inherit $\boldsymbol{W}_Q$, $\boldsymbol{W}_K$, $\boldsymbol{W}_V$, $\boldsymbol{W}_O$ parameters (blue) from trained Transformer layer and carefully initialize the extra parameters (orange) including $a$, $\boldsymbol{W}_\gamma$, $\boldsymbol{W}_{\text{conv}}$, and $\boldsymbol{W}_G$ in Mamba-2 to initially mimic the Transformer's behavior, providing a strong foundation for subsequent distillation.
    }
    \vspace{-0.3cm}
    \label{fig:Mamba-2_from_transformer}
\end{figure}

To transfer as much knowledge as possible from quadratic Transformer to linear Mamba-2, we initialize Mamba-2 from Transformer at each layer.
By comparing Eq.~\ref{eq:attn} and Eq.~\ref{eq:Mamba-2}, we can find that Mamba-2 shares the similarity with Transformer, which means we can directly inherit $\boldsymbol{W}_Q, \boldsymbol{W}_K, \boldsymbol{W}_V$ and $\boldsymbol{W}_O$ projection parameters at each layer instead of building from scratch. 
Furthermore, we need to introduce extra parameters $\boldsymbol{W}_\gamma$ and $a$ for state space modeling, replacing the attention mechanism.
For better replacement and ease the training difficulty~\cite{trockman2024mimetic}, we initialize $\boldsymbol{W}_\gamma$ and $a$ to make the gating term $\gamma_t$ close to $1$ at the beginning of training, which means we begin by memorizing all historical information without selectivity.

Beyond the core SSM mechanism, we also introduce extra causal convolution and output gating for enhanced positional awareness and expressiveness.
To eliminate the initial impact of causal convolution, we initialize the weights and biases to make it function as an identity layer (\ie, the output of causal convolution is the same as the input) without affecting the original function of SSM at the beginning of training.

The other parts of the model such as the MLP layers and text and image patch embedding layers are directly inherited from the original Transformer-based VLM and kept as frozen.

\begin{figure*}[t]
    \centering
    \includegraphics[width=1.0\linewidth]{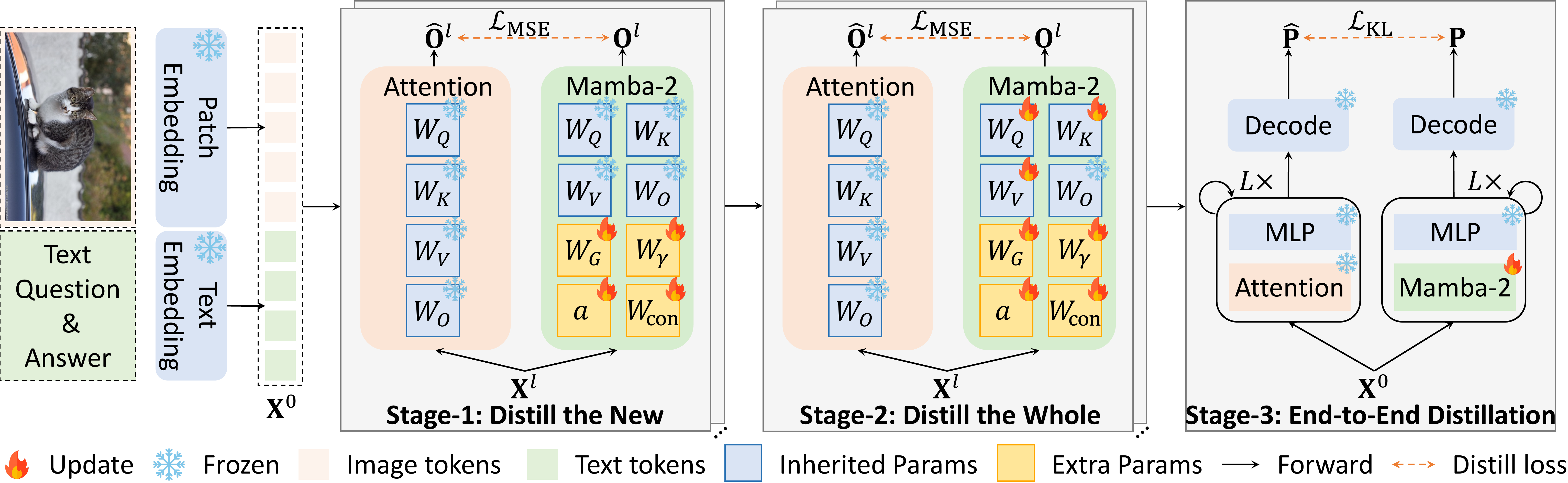}
    \caption{\textbf{Progressive distillation pipeline of our \name{}.} We keep MLP layers, text and image patch embedding layers and freeze them in subsequent distillation training stages. Stage-1: Train the newly-introduced SSM-specific parameters while freezing inherited Transformer parameters in a layer-wise manner. Stage-2: Train all parameters to align Mamba's state representation with Transformer in a layer-wise manner. Stage-3: Train all the Mamba layers of the model to align the end-to-end behavior with the teacher Transformer-based VLM.
    }
    \label{fig:distill_pipeline}
\end{figure*}

\subsection{Stage-1: Layerwise Distillation for the Newly Introduced SSM Parameters}
\label{sec:stage1}

We first perform layerwise distillation for the introduced extra parameters to align the proposed Mamba-2 layer with the original trained Transformer layer.
Specifically, we instantiate the trained Transformer-based VLM as teacher model, and the transferred Mamba-2 VLM model as student model.
The only difference lies in the sequence mixer layer. We feed the multimodal sequence into the teacher model.

To keep the layerwise alignment and diminish the accumulated error of the cascading layers, we input the $i$-th Mamba-2 layer with the output of the $i-1$-th Transformer layer, i.e., $i$-th Mamba-2 layer and $i$-th Transformer layer have the same input. And we align the layerwise behavior by applying the MSE distillation loss between the output of the $i$-th Mamba-2 layer and the output of the $i$-th Transformer layer:
\begin{align}
    \begin{split}
    \phi_{\text{stage1}}^i &= \{a^i, \boldsymbol{W}_{\gamma}^i, \boldsymbol{W}_{\text{conv}}^i, \boldsymbol{W}_{G}^i\}, \\
    \min_{\{\phi_{\text{stage1}}^i\}_{i=1}^{L}} &\sum_{i=1}^{L}\mathcal{L}_{\mathrm{MSE}}(\text{Attn}(\mathbf{X}^{i}), \text{Mamba-2}_{\phi_{\text{stage1}}^i}(\mathbf{X}^{i})),
\end{split}
\end{align}
where $\phi_{\text{stage1}}^i$ is the trainable parameters of the $i$-th Mamba-2 layer, which only includes the introduced extra parameters $a^i, \boldsymbol{W}_{\gamma}^i, \boldsymbol{W}_{\text{conv}}^i, \boldsymbol{W}_{G}^i$.
$\mathbf{X}^{i}$ is the input sequence to the $i$-th Mamba-2 layer and Transformer layer, $\text{Attn}(\mathbf{X}^{i})$ is the output of the $i$-th teacher Transformer layer, $\text{Mamba-2}_{\phi_{\text{stage1}}^i}(\mathbf{X}^{i})$ is the output of the $i$-th student Mamba-2 layer.

\subsection{Stage-2: Layerwise Distillation for the Whole Mamba-2 Parameters}
\label{sec:stage2}
After the stage-1 distillation, we have obtained good initialization of the introduced extra parameters, and we further train all the Mamba-2 parameters to better align the layerwise behavior of the student Mamba-2 with the teacher Transformer.
The only difference between the stage-1 and stage-2 is that we further include the parameters of $\boldsymbol{W}_Q, \boldsymbol{W}_K, \boldsymbol{W}_V$ for optimizing the distillation loss:
\begin{align}
    \begin{split}
    &\phi_{\text{Stage2}}^i = \{a^i, \boldsymbol{W}_{\gamma}^i, \boldsymbol{W}_{\text{conv}}^i, \boldsymbol{W}_{G}^i, \boldsymbol{W}_Q^i, \boldsymbol{W}_K^i, \boldsymbol{W}_V^i\}, \\
    &\min_{\{\phi_{\text{Stage2}}^i\}_{i=1}^{L}} \sum_{i=1}^{L}\mathcal{L}_{\mathrm{MSE}}(\text{Attn}(\mathbf{X}^{i}), \text{Mamba-2}_{\phi_{\text{Stage2}}^i}(\mathbf{X}^{i})),
\end{split}
\end{align}

\subsection{Stage-3: End-to-End Distillation}
\label{sec:stage3}
Beyond the layerwise alignment, the final stage-3 distillation aims to align the end-to-end behavior of the student Mamba-2 with the teacher Transformer.
Specifically, we input the same multi-modal sequence to both the teacher Transformer and the student Mamba-2 without sharing the intermediate output.
For the output of the teacher model and the student model, we apply the word-level KL-Divergence loss, in other words, they are used as soft labels, we enforce the output logits of the student model to be close to the output logits of the teacher model:
\begin{align}
\begin{split}
    &\phi_{\text{Stage-3}} = \{a^i, \boldsymbol{W}_{\gamma}^i, \boldsymbol{W}_{\text{conv}}^i, \boldsymbol{W}_{G}^i, \boldsymbol{W}_Q^i, \boldsymbol{W}_K^i, \boldsymbol{W}_V^i\}_{i=1}^{L}, \\
    &\min_{\phi_{\text{Stage-3}}} \mathcal{L}_{\mathrm{KL}}(\text{Teacher-}(\mathbf{X}^0), \text{Student}_{\phi_{\text{Stage3}}}(\mathbf{X}^{0})),
\end{split}
\end{align}
where $\mathbf{X}^0$ is the same multi-modal input sequence to the teacher model and the student model, $\phi_{\text{Stage3}}$ is the trainable parameters of the student model.

\subsection{Architecture}
\label{sec:arch}
Our \name{} builds upon HoVLE, a decoder-only VLM that consists of 32 Transformer layers. 
For \name{}-linear, we convert each Transformer layer into a Mamba-2 layer while preserving the MLP layers, resulting in a linear-complexity decoder-only VLM. To enhance model expressiveness, we adopt a multi-head design in our Mamba-2 layers by partitioning the SSM into multiple groups and implementing shared queries across groups, consistent with the grouped query attention used in HoVLE.

For \name{}-hybrid, we introduce a systematic layer conversion scheme. Specifically, within every fixed number of consecutive layers, we preserve the first layer as Transformer and convert the remaining layers to Mamba-2. This hybrid scheme maintains the Transformer's modeling capacity at critical feature hierarchies while leveraging Mamba-2's linear complexity for the majority of computation. Such design enables an effective and flexible trade-off between computational efficiency and model capability, suitable for various deployment scenarios with varied requirements. In this paper, we set the interval as 4, building \name{}-hybrid with 8 Transformer layers and 24 Mamba-2 layers in total.

%% file: secs/4_exp.tex
\input{tables/main_results}
\input{tables/ablation_efficiency}

\section{Experiment}
\subsection{Implementation Detail}

\boldparagraph{Training.}
All models are trained using 8 NVIDIA A800 80GB GPUs with BF16 precision and DeepSpeed ZeRO-2~\cite{rajbhandari2020zero,rasley2020deepspeed}. The distillation process utilizes SOLO's~\cite{solo} supervised fine-tuning dataset, comprising 1.7M samples across both language-only and image-text paired instances. We employ the AdamW~\cite{adamw} optimizer with $\beta=(0.9,0.999)$, gradient clipping at 5.0, and a Warmup-Stable-Decay (WSD) scheduler with 10\% warmup and 10\% decay periods. For stages-1 and stage-2 distillation, we use a batch size of 128, train for 20K steps, and set weight decay to 0.05, with learning rates of $1 \times 10^{-3}$ and $5 \times 10^{-4}$ respectively. Stage-3 distillation employs a reduced batch size of 64, continues for 20K steps with weight decay at 0.05, and uses a learning rate of $5 \times 10^{-5}$.

\boldparagraph{Evaluation benchmarks.}
We evaluate our model on 9 diverse public benchmarks, encompassing 6 general VLM benchmarks and 3 visual question answering tasks. 
The general VLM benchmarks include: MME~\cite{mme}, which evaluates visual perception and reasoning through true/false questions; 
MMBench~\cite{mmb}, which assesses model robustness through multiple-choice questions; 
POPE~\cite{pope}, which evaluates object hallucination; 
SEED~\cite{seed}, which gauges open-world multi-modal understanding; 
MMMU~\cite{mmmu}, which scrutinizes models with college-level multi-discipline reasoning tasks;
and MM-Vet~\cite{mmvet}, which evaluates the model on 16 emergent tasks from core visual and linguistic capabilities.
The visual question answering benchmarks comprise: TextVQA~\cite{textvqa}, which evaluates optical character recognition (OCR) capabilities and text-based reasoning; 
ScienceQA~\cite{sqa}, which tests scientific image comprehension; 
and GQA~\cite{gqa}, which assesses real-world visual reasoning and compositional question answering.

For the specific score in the comparison, we report the MME-perception score as the MME score, MMB score is calculated on the MMBench-EN split, and the POPE score is calculated by averaging across its three categories.

\subsection{Main Comparison}
In Tab.~\ref{tab:results_general}, we compare \name{} with previous encoder-based and decoder-only VLMs on 9 popular benchmarks. We highlight the following findings:
\begin{itemize}
\item \name{} only performs distillation as the training recipe, which requires much lower training cost in two aspects: (1) dataset collection--unlike other methods require separate curated datasets for pre-training (PT) and supervised fine-tuning (SFT), our distillation recipe only needs a single SFT dataset; (2) trainable parameters--our method only updates 14.7\% parameters for \name{}-linear and 11.2\% parameters for \name{}-hybrid during training, while other methods require training most of parameters.
\item  \name{}-linear surpasses previous SOTA Transformer-based decoder-only VLM EVE-7B on 6/9 benchmarks (\ie, MME, MMB, POPE, SEED, MM-Vet, ScienceQA), while matching the performance on the left 3 benchmarks with 2$\times$ fewer parameters.
Even compared with encoder-based VLMs (\eg, MobileVLM-3B, LLaVA-phi), \name{}-linear still demonstrates a comparable performance, while the computation complexity is reduced to linear complexity.
\item \name{}-linear matches the performance of recent encoder-based linear-complexity VLMs (VisualRWKV-3B and VL-Mamba-3B), while significantly outperforming them on the ScienceQA benchmark.
\item By interleaving with the Transformer layers, \name{}-hybrid achieves improved performance on all benchmarks over \name{}-linear, significantly narrowing the gap with the teacher Transformer-based  HoVLE and outperforming the linear complexity encoder-based VLMs (VisualRWKV-3B and VL-Mamba-3B).
\end{itemize}

\subsection{Efficiency Analysis}

\boldparagraph{Fixed prompt and fixed decode length.}
We directly follow the benchmark recipe of Cobra in Tab.~\ref{tab:efficiency_comp}, where we prompt the VLM model with the same example image and question ``Describe the image specifically'', and set the number of output tokens to 256.
We record the total time of the VLM model, which includes the image/text prompt prefilling time and the decoding time. The speed (tokens/s) is calculated by the number of output tokens (\ie, 256) divided by the total time.
We compare our method with 3 transformer-based VLMs and 2 linear-complexity encoder-based VLMs of similar parameter scale. All the evaluations are conducted on the same single NVIDIA RTX 4090 GPU.

Thanks to the fixed hidden state rooted in linear-complexity modeling, \name{}-linear/hybrid achieve nearly 4$\times$ faster inference speed than all the Transformer-based VLMs.
\name{}-linear/hybrid also outperforms the linear complexity encoder-based VLMs (Cobra-3.5B and VisualRWKV-3B) by a large margin (about 30 tokens/s and 3$\times$ faster, respectively) due to the simple decoder-only architecture.

\input{tables/ablation_stage}

\input{tables/ablation_init}

\boldparagraph{Increasing context length.}
Long context processing has emerged as a crucial capability in modern VLMs, becoming increasingly important for high-resolution image/video understanding~\cite{videollm, hrvlm} and long chain-of-thought multimodal reasoning~\cite{llavacot,Lightman2023LetsVS,deepseekr1, s1, k1p5}, which often require processing sequences of thousands of tokens.
To showcase the efficiency of the proposed \name{} in this application, we compare our model with Transformer-based HoVLE in the same single NVIDIA RTX 4090 GPU. We report the GPU memory usage and the latency of the model for decoding the next token.

As shown in Fig.~\ref{fig:teaser}, thanks to the efficient implementation of FlashAttention2, HoVLE demonstrates stable and low latency under 4K token length.
As the context token length reaches 8K and beyond, the latency and memory of HoVLE increase linearly with the token length due to the growing Key-Value cache, when the token length reaches 128K, HoVLE squeezes out of the GPU memory and fails to decode.
On the contrary, \name{}-linear exhibits low and stable latency and memory usage with increasing token length, and the inference cost of \name{}-hybrid increases much slower than HoVLE, which can still decode at the 128K token length.
Specifically, at the 103K token length, \name{}-linear demonstrates 20.6$\times$ speedup compared to HoVLE and saves 75.8\% GPU memory, while \name{}-hybrid achieves 13.5$\times$ speedup and saves 60.2\% GPU memory.

\subsection{Ablation Study}

\input{tables/ablation_hybrid}

\input{tables/ablation_hybrid_strategy}

\boldparagraph{Stage importance.}
As shown in Tab.~\ref{tab:stage}, direct weight transfer from Transformer to Mamba-2 without distillation (Sec.~\ref{sec:init}) lost the multi-modal understanding ability.
By progressively adding the designed distillation stages, the model's performance is increasingly improved.
When comparing ID-7 and ID-8, we can see that the proposed extra parameter distillation stage-1 decouples the optimization and eases the training, leading to a better alignment, with consistent improvements across all metrics 
(48 in MME, 1.7 in POPE, 6.6 in TextVQA, 7.1 in ScienceQA).

\boldparagraph{Parameter initialization.}
In Tab.~\ref{tab:init}, 
We compare with the ``from scratch'' strategy used in Phi-Mamba~\cite{bick2024transformers}, which replace the trained Transformer layer with directly initialized Mamba-2 layer without inheriting the parameters,
and the ``inherit $\boldsymbol{W}_{Q,K,V}$'' strategy used in LoLCATs~\cite{zhang2024lolcats} and Mamba in the LLaMA~\cite{wang2024mamba}, which exploit the similarity and only inherit the parameters of $\boldsymbol{W}_{Q,K,V,O}$ from Transformer to Mamba-2.
The results validate the superiority of our proposed parameter initialization strategy, which should not only inherit the trained parameters but also initialize the extra introduced parameters by mimicking the original attention mechanism.

\boldparagraph{Hybrid architecture.}
The proposed distillation recipe is more flexible than the previous training recipe used in building linear-complexity encoder-based VLM, which requires the trained linear-complexity LLM and can not modify the architecture.
As shown in Tab.~\ref{tab:hybrid}, we can build a hybrid architecture with varied interleaved Transformer layers, enabling the flexible trade-off between performance and efficiency.
By increasing the number of Transformer layers, the performance is gradually improved.
The hybrid architecture with 24 Mamba-2 layers and 8 Transformer layers can achieve comparable performance with a minor decrease compared to the full Transformer model HoVLE.

\boldparagraph{Hybrid strategy.}
In Tab.~\ref{tab:hybrid_strategy}, we explore specific hybrid strategies while fixing the number of interleaved Transformer layers to 8. 
We study 4 interleaving strategies: 
(1) Tail-stacked: stacking all 8 Transformer layers at the top of the network. 
(2) Head-stacked: stacking all 8 Transformer layers at the bottom of the network;
(3) Tail-interleaved: interleaving a Transformer layer at the tail of every 4-layer block; 
(4) Head-interleaved: interleaving a Transformer layer at the head of every 4-layer block; 
The results demonstrate that the Head-interleaved strategy is the most effective, achieving the best performance across all metrics

%% file: tables/main_results.tex
\begin{table*}[t!]
    \centering
    \small
    
    \scalebox{0.90}{
    \setlength{\tabcolsep}{1.0pt}
    \begin{tabular}{l c c c r | c c c c c c |c  c c }
    \toprule
    \multirow{1}{*}{Method} & \multirow{1}{*}{Recipe} & \multirow{1}{*}{Complexity} & \multirow{1}{*}{\# P.} & \multirow{1}{*}{\# T.P.}& MME & MMB &POPE & \multicolumn{1}{c} {SEED} & MMMU & MM-Vet& TQA & SQA-I  & \multicolumn{1}{c}{GQA} \\
    \midrule
    \rowcolor{gray!14}
    \multicolumn{14}{l}{\textbf{\textit{Encoder-based VLMs}}} \\ 
    OpenFlamingo~\cite{openflamingo} & \underline{PT, SFT}& Quadratic & 9B& 96.6\%  & - & 4.6 & - & - & - & - & 33.6 & - & - \\
    MiniGPT-4~\cite{minigpt} & \underline{PT, SFT}& Quadratic & 13B& 94.8\%  & 581.7 & 23.0 & - & - & -& 22.1 & - & - & 32.2  \\
    Qwen-VL~\cite{qwenvl} & \underline{PT, SFT}& Quadratic & 7B& 100.0\%  & - & 38.2 & - & 56.3 & - & - & 63.8 & 67.1 & 59.3\\ 
    LLaVA-Phi~\cite{llavaphi}  & \underline{PT, SFT}& Quadratic & 3B& 90.0\%  & 1335.1 & 59.8 & 85.0 & - & - & 28.9& 48.6 & 68.4 & - \\
    MobileVLM-3B~\cite{mobilevlm} & \underline{PT, SFT}& Quadratic & 3B& 90.0\%  & 1288.9 & 59.6 & 84.9 & - & - & - & 47.5 & 61.0 & 59.0  \\
    VisualRWKV~\cite{visualrwkv} & \underline{PT, SFT}&  \textbf{Linear} & 3B& 90.0\%  & 1369.2 & 59.5 & 83.1 & - & - & - & 48.7 & 65.3 & 59.6 \\
    VL-Mamba~\cite{vlmamba} & \underline{PT, SFT}&  \textbf{Linear} & 3B& 90.0\%  & 1369.6 & 57.0 & 84.4 & - & -& 32.6 & 48.9 & 65.4 & 56.2 \\
    Cobra~\cite{cobra} & \underline{PT, SFT}&  \textbf{Linear} & 3.5B& 82.6\%  & - & - & \textbf{88.4} & - & - & - & 58.2 & - & \textbf{62.3}\\
    \midrule
    \rowcolor{gray!14}
    \multicolumn{14}{l}{\textbf{\textit{Decoder-only VLMs}}} \\
    Fuyu-8B (HD)~\cite{fuyu} & \underline{PT, SFT}& Quadratic & 8B& 100.0\%  & 728.6 & 10.7 & 74.1 & - & - & 21.4 & - & - & -\\
    SOLO~\cite{solo} & \underline{PT, SFT}& Quadratic &  7B& 100.0\%   & 1001.3 & - & - & 64.4 & - & - & - & 73.3 & -   \\    
    Chameleon-7B~\cite{chameleon}  & \underline{PT, SFT}& Quadratic &  7B& 100.0\%   & 170 & 31.1 & - & 30.6 & 25.4 & 8.3 & 4.8 & 47.2 & -\\  
    EVE-7B~\cite{eve}  & \underline{PT, SFT}& Quadratic &  7B& 100.0\%  & 1217.3 & 49.5 & 83.6 & 61.3 & \underline{32.3} & 25.6& 51.9 & 63.0 & 60.8 \\
    Emu3~\cite{emu3} & \underline{PT, SFT}& Quadratic & 8B& 100.0\%  & - & 58.5 & 85.2 & \underline{68.2} & 31.6 & \underline{37.2} & \underline{64.7} & \underline{89.2} & 60.3\\
    HoVLE~\cite{hovle} & DT, PT, SFT & Quadratic & \textbf{2.6B}& 100.0\%  & \textbf{1433.5} & \textbf{71.9} & \underline{87.6} & \textbf{70.7} & \textbf{33.7} & \textbf{44.3} & \textbf{66.0} & \textbf{94.8} & \underline{60.9} \\
    \rowcolor{green!15}
    \name{} & \textbf{DT} & \textbf{Linear} & \underline{2.7B}& \underline{14.7\%}  &1303.5 & 57.2 & 85.2 & 62.9& 30.7  & 31.1 &47.7 & 79.2 & 57.4 \\
    \rowcolor{yellow!15}
    \name{} & \textbf{DT} & \underline{Hybrid} & \underline{2.7B}& \textbf{11.2\%}  & \underline{1371.1} & \underline{63.7} & 86.7 & 66.3 & \underline{32.3} & 36.9 & 55.1 & 86.9 & 59.3  \\
    
    \bottomrule
    \end{tabular}
    }
    \vspace{-1em}
    \caption{\textbf{Comparison with existing VLMs on general VLM benchmarks.} ``Recipe'' denotes the adopted training recipe. ``PT'', ``SFT'', and ``DT'' denote the pre-training, supervised fine-tuning, and distillation training, respectively. ``Complexity'' denotes the model computation complexity with respect to the number of tokens. ``\# P.'' denotes the number of total parameters. ``\# T.P.'' denotes the percentage of trainable parameters ($\frac{\text{trainable paramters}}{\text{total parameters}}$). The best performance is highlighted in \textbf{bold} and the second-best result is \underline{underlined}.}
    \label{tab:results_general}
    \end{table*}

%% file: tables/ablation_efficiency.tex
\begin{table*}[]
    \centering
    \small
    \scalebox{0.95}{
    \setlength{\tabcolsep}{1pt}
    \begin{tabular}{c | c c c| c c |c c}
    \toprule
    Model & LLM Backbone & Vision Encoder & Total Params & Visual Tokens & Output Tokens & Speed (tokens/s) & Total (s)\\
    \midrule
    LLaVA-Phi & Phi2-2.7B&  CLIP ViT-L/14 &3.1B &576 & 256& 26.92& 9.51\\
    MobileVLM-3B & LLaMA-2.7B &  CLIP ViT-L/14&3.1B& 144 & 256 & 35.26 & 7.26  \\
    HoVLE & \multicolumn{2}{c}{32-layer Transformer} &\textbf{2.6B}&768&256&33.03&7.75\\
    \midrule
    Cobra-3.5B & Mamba-2.8B & DINOv2 + SigLIP ViT-SO &3.5B&729&256&99.22&2.58\\
    VisualRWKV-3B & RWKV6-3B &CLIP ViT-L/14 &3.4B&577&256&41.34&6.19\\
    \rowcolor{green!15}
    \name{}-linear & \multicolumn{2}{c}{32-layer Mamba2}&\underline{2.7B}&768&256&\underline{132.43}&\underline{1.93}\\
    \rowcolor{yellow!15}
    \name{}-hybrid & \multicolumn{2}{c}{24-layer Mamba2 + 8-layer Transformer }&\underline{2.7B}&768&256&\textbf{134.77}&\textbf{1.90}\\
    \bottomrule
    \end{tabular}
    }
    \vspace{-1em}
    \caption{\textbf{Inference efficiency comparison under same multimodal prompt and fixed decode length.} We compare with VLMs of the similar parameter scale (3B) across encoder-based, decoder-only, quadratic-complexity, and linear-complexity. The results highlight the speed advantage of \name{}-linear/hybrid. The benchmark recipe directly follows Cobra, and we report the results on the same single NVIDIA RTX 4090 GPU. Note that ``Total Time'' includes the time of both prefilling and decoding, and ``Speed'' = ``Output Tokens'' / ``Total Time''.}
    \vspace{-1em}

    \label{tab:efficiency_comp}
 \end{table*}

%% file: tables/ablation_stage.tex
\begin{table}[t!]
    \centering
    \small
    \setlength{\tabcolsep}{2pt}
    \begin{tabular}{l|c c c | c  c c c}
    \toprule
       ID &  Stage1 & Stage2 & Stage3 &  MME &  POPE & TextVQA & SQA-I \\
    \midrule
    1&& & & NAN & NAN & NAN & NAN \\
    2&\checkmark & & & 969.8 & 70.6 & 13.47 & 40.8 \\
    3&   & \checkmark &  & 1007.1 & 72.9 & 25.5 & 52.1 \\
    4& & & \checkmark       & 1188.4 & 83.0 & 40.0 & 63.4  \\
    5&    \checkmark & \checkmark &        & 1108.9 & 75.3 & 28.0 & 59.3 \\
    6&    \checkmark & & \checkmark & 1263.1 & 84.0 & 42.5 & 77.1 \\
    7&    & \checkmark & \checkmark & 1255.5 & 83.5 & 41.1  & 72.1 \\
    \rowcolor{green!15}
    8&   \checkmark & \checkmark & \checkmark & 1303.5 & 85.2 & 47.7 & 79.2 \\
    \bottomrule
    \end{tabular}
    \vspace{-1em}
    \caption{\textbf{Ablation for training stages.}}
    \vspace{-1em}
    \label{tab:stage}
 \end{table}

%% file: tables/ablation_init.tex
\begin{table}[t!]
    \centering
    \small
    \setlength{\tabcolsep}{3pt}
    \begin{tabular}{c | c c c c}
    \toprule
    Init Strategy & MME &  POPE & TextVQA & SQA-I \\
    \midrule
    from scratch & 1214.0 & 83.1 & 40.0 & 67.4 \\
    inherit $\boldsymbol{W}_{Q,K,V}$ & 1222.6 & 84.0 & 41.9 & 73.3 \\
    \rowcolor{green!15}
    inherit $\boldsymbol{W}_{Q,K,V}$ + mimic&  1303.5 & 85.2 & 47.7 & 79.2 \\
    \bottomrule
    \end{tabular}
    \vspace{-1em}
    \caption{\textbf{Ablation for parameter initialization.} }
    \vspace{-1em}

    \label{tab:init}
 \end{table}

%% file: tables/ablation_hybrid.tex
\begin{table}[t!]
    \centering
    \small
    \begin{tabular}{c | c  c c c}
    \toprule
    Attention Layers & MME &  POPE & TextVQA & SQA-I \\
    \midrule
    \rowcolor{green!15}
    0 &1303.5 & 85.2 & 47.7 & 79.2 \\
    \midrule
    1 & 1304.3& 85.5 & 48.0 & 79.3 \\
    2 &  1318.4 & 86.3 & 48.4 & 79.9 \\
    4 & 1329.1 & 86.8 & 51.5 & 82.8 \\
    \rowcolor{yellow!15}
    8 &1371.1 & 86.7 & 55.1 & 86.9 \\
    \midrule
    32 & 1433.5 & 87.6 & 66.0 & 94.8 \\
    \bottomrule
    \end{tabular}
    \vspace{-1em}
    \caption{\textbf{Ablation for the number of interleaved attention layers.} ``0'' denotes \name{}-pure, ``8'' denotes \name{}-hybrid, ``32'' denotes the full Transformer model HoVLE.}
    \vspace{-1em}
    \label{tab:hybrid}
 \end{table}

%% file: tables/ablation_hybrid_strategy.tex
\begin{table}[t!]
    \centering
    \small
    \begin{tabular}{c | c c c c c}
    \toprule
    Hybrid strategy &MME &  POPE & TextVQA & SQA-I\\
    \midrule
    Tail-stacked & 1305.5 & 85.9 & 53.7 & 79.4 \\
    Head-stacked & 1329.4 & 85.9 & 55.0 & 80.8  \\
    Tail-interleaved & 1308.3 & 86.1 & 55.0 & 86.5 \\
    \rowcolor{yellow!15}
    Head-interleaved & 1371.1 & 86.7 & 55.1 & 86.9 \\
    \bottomrule
    \end{tabular}
    \vspace{-1em}
    \caption{\textbf{Ablation for hybrid strategy.}}
    \vspace{-1em}
    \label{tab:hybrid_strategy}
 \end{table}

%% file: secs/5_conclusion.tex
\section{Conclusion}
We presented \name{}, a novel framework for building linear-complexity decoder-only VLMs with only moderate academic resources through the proposed distillation recipe, eliminating the need for pre-trained linear-complexity LLMs and vision encoders. Our recipe enables both pure linear and hybrid architectures, achieving competitive performance while significantly reducing computational costs. Experimental results demonstrate that \name{}-linear matches or exceeds the performance of existing linear-complexity and quadratic-complexity VLMs, while \name{}-hybrid further improves performance through flexible efficiency-performance trade-offs with interleaved Transformer layers. At 103K tokens, \name{}-linear achieves up to 20.6× speedup and 75.8\% memory reduction compared to Transformer-based teacher HoVLE, while \name{}-hybrid achieves 13.5$\times$ speedup and saves 60.2\% GPU memory. These results validate the effectiveness of our distillation recipe for building linear-complexity decoder-only VLMs suitable for long-context applications.